\documentclass[letterpaper]{article} % DO NOT CHANGE THIS
\usepackage{aaai2026}  % DO NOT CHANGE THIS
\usepackage{times}  % DO NOT CHANGE THIS
\usepackage{helvet}  % DO NOT CHANGE THIS
\usepackage{courier}  % DO NOT CHANGE THIS
\usepackage[hyphens]{url}  % DO NOT CHANGE THIS
\usepackage{graphicx} % DO NOT CHANGE THIS
\usepackage{latexsym}
\usepackage{amssymb}
\usepackage{amsmath}
\usepackage{booktabs}
\usepackage{multirow}
\usepackage{colortbl}
\usepackage{makecell}
\usepackage{natbib}  % DO NOT CHANGE THIS AND DO NOT ADD ANY OPTIONS TO IT
\usepackage{caption} % DO NOT CHANGE THIS AND DO NOT ADD ANY OPTIONS TO IT
\usepackage[ruled,linesnumbered]{algorithm2e}
\usepackage{float}
\usepackage{listings}
\newfloat{listing}{tb}{lst}[section]
\floatname{listing}{Listing}
\DeclareCaptionStyle{ruled}{labelfont=normalfont,labelsep=colon,strut=off}
\title{Improving Generalization in LLM Structured Pruning \\ via Function-Aware Neuron Grouping}
\author{
    Tao Yu\textsuperscript{\rm 1 \rm 2},
    Yongqi An\textsuperscript{\rm 1 \rm 2},
    Kuan Zhu\textsuperscript{\rm 1},
    Guibo Zhu\textsuperscript{\rm 1 \rm 2 \rm 3},
    Ming Tang\textsuperscript{\rm 1 \rm 2},
    Jinqiao Wang\textsuperscript{\rm 1 \rm 2 \rm 3}\thanks{Corresponding author.}
}
\affiliations{
    \textsuperscript{\rm 1}Foundation Model Research Center, Institute of Automation, Chinese Academy of Sciences, Beijing, China\\
    \textsuperscript{\rm 2}School of Artificial Intelligence, University of Chinese Academy of Sciences, Beijing, China\\
    \textsuperscript{\rm 3}Wuhan AI Research, Wuhan, China\\
    yutao2022@ia.ac.cn, 
    \{yongqi.an, kuan.zhu, gbzhu, tangm, jqwang\}@nlpr.ia.ac.cn
}

\begin{document}

\maketitle

\begin{abstract}
Large Language Models (LLMs) demonstrate impressive performance across natural language tasks but incur substantial computational and storage costs due to their scale. Post-training structured pruning offers an efficient solution. However, when few-shot calibration sets fail to adequately reflect the pretraining data distribution, existing methods exhibit limited generalization to downstream tasks. To address this issue, we propose Function-Aware Neuron Grouping (FANG), a post-training pruning framework that alleviates calibration bias by identifying and preserving neurons critical to specific function. FANG groups neurons with similar function based on the type of semantic context they process and prunes each group independently. During importance estimation within each group, tokens that strongly correlate with the functional role of the neuron group are given higher weighting. Additionally, FANG also preserves neurons that contribute across multiple context types. To achieve a better trade-off between sparsity and performance, it allocates sparsity to each block adaptively based on its functional complexity. Experiments show that FANG improves downstream accuracy while preserving language modeling performance. It achieves the state-of-the-art (SOTA) results when combined with FLAP and OBC, two representative pruning methods. Specifically, FANG outperforms FLAP and OBC by 1.5\%–8.5\% in average accuracy under 30\% and 40\% sparsity.
\end{abstract}

\section{Introduction}

In recent years, Large Language Models (LLMs) \cite{gpt3,llama} have shown remarkable capabilities in generating high-quality text and tackling a wide range of downstream tasks. Despite their powerful performance, these models typically contain vast numbers of parameters, resulting in increased computational and storage demands that hinder efficient deployments \cite{effsurvey}. Structured pruning has emerged as an effective solution to these challenges by removing redundant neurons or attention heads \cite{shearedllama,ziplm}, thereby reducing both computation and memory costs. 

\begin{figure}[t]
\centering
\includegraphics[width=0.4\textwidth]{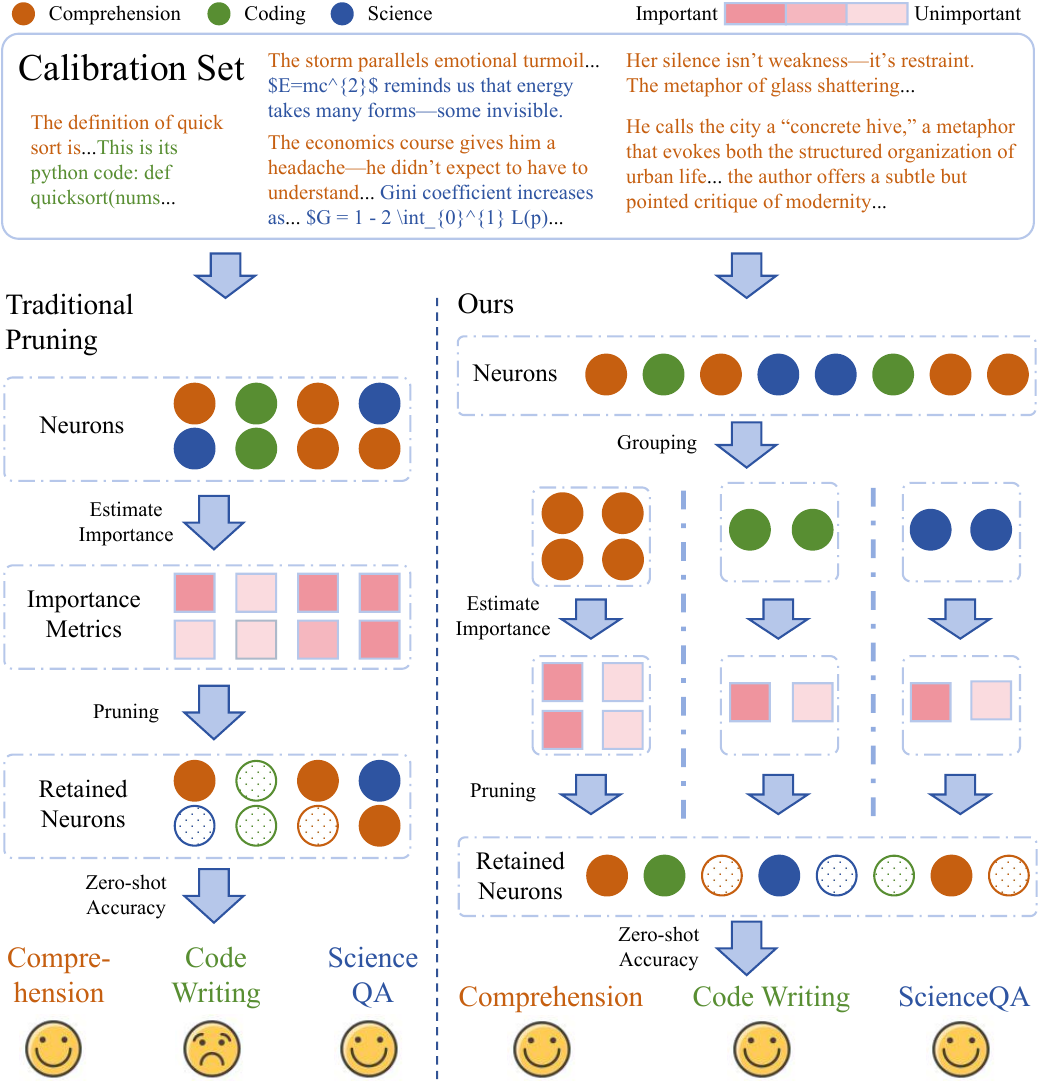} % Reduce the figure size so that it is slightly narrower than the column. Don't use precise values for figure width.This setup will avoid overfull boxes.
\caption{Compared to traditional post-training pruning, our method mitigates calibration set distribution mismatch by pruning neurons based on functional grouping, reducing the risk of misestimating important neurons and improving generalization.}
\label{fig:intro}
\end{figure}

Given the massive scale of LLMs, traditional prune-and-retrain methods are often computationally prohibitive \cite{shearedllama, darwinlm}. A more practical alternative is post-training pruning without fine-tuning \cite{flap,modegpt,sobp,slimgpt}, which eliminates redundant parameters in a one-shot manner while employing numerical compensation to recover performance. Specifically, these methods compute importance metrics for neurons based on a calibration set, rank the neurons, and remove those with the lowest importance. The calibration set is typically sampled from pretraining data to preserve modeling ability on the pretraining task, which is commonly measured by perplexity. It also provides a certain degree of generalization, as reflected in zero-shot accuracy on downstream tasks. In practice, these methods remain effective even with a limited number of calibration samples.

However, despite its effectiveness, the small size of the calibration set makes the model prone to overfitting. When the calibration data fails to adequately reflect the distribution of the pretraining corpus, this overfitting can lead to degraded generalization \cite{truth}. In particular, the importance of functionally critical neurons may be misestimated, resulting in their erroneous removal and irreversible accuracy loss on downstream tasks, as shown in Fig.~\ref{fig:intro}. This raises a key question:
\textbf{\textit{Can we enhance generalization ability while preserving the efficiency of pruning and maintaining performance on the pretraining task?}}

To address this question, we draw inspiration from interpretability research \cite{xai_neuron,xai_probe,xai_knowledge,xai_dm}, which suggests that LLMs exhibit functional specialization—analogous to the human brain—where different neurons are responsible for processing distinct types of contextual information. Building on this insight,  we propose \textbf{\textit{Function-Aware Neuron Grouping (FANG)}}, a post-training structured pruning framework designed to retain functional diversity and improve generalization. 

FANG consists of three core components: (1) A \textbf{\textit{function-aware pruning strategy}}, in which neurons are grouped by their functional roles and pruned independently. During importance estimation within each group, tokens most semantically aligned with the group’s function are assigned greater weight. (2) A \textbf{\textit{shared neuron group retention}} mechanism that identifies and preserves neurons contributing across multiple context types. (3) An \textbf{\textit{adaptive sparsity allocation}} scheme that assigns lower sparsity to functionally complex blocks based on a complexity metric. These designs retain both specialized and general capacities, support more balanced pruning, and improve generalization across various downstream tasks.

\begin{figure*}[t]
\centering
\includegraphics[width=0.9\textwidth]{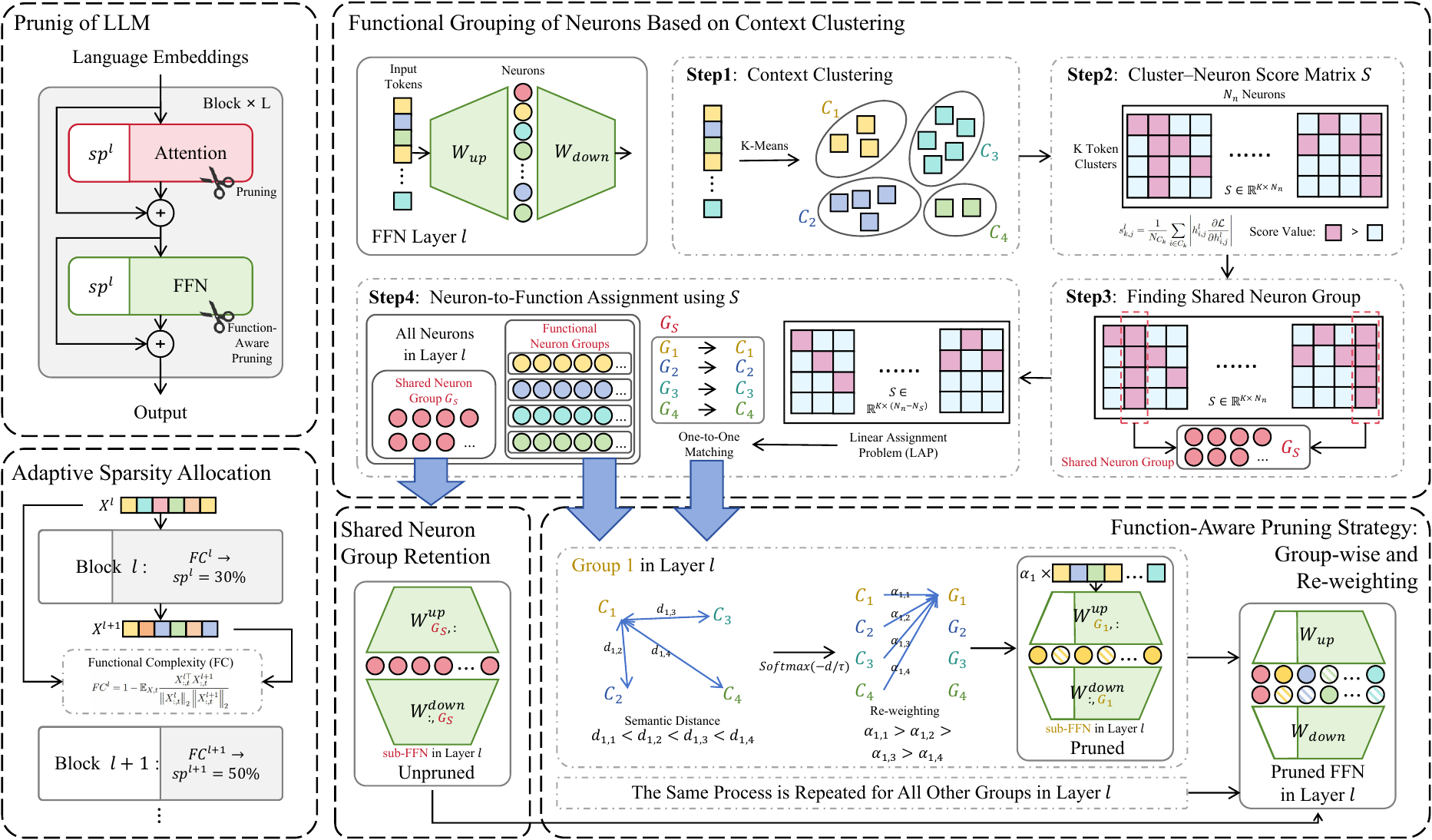} % Reduce the figure size so that it is slightly narrower than the column. Don't use precise values for figure width.This setup will avoid overfull boxes.
\caption{Overview of our Function-Aware Neuron Grouping (FANG). For attention heads, we adopt baseline pruning methods such as OBC and FLAP. For FFN layers, neurons are grouped based on their functional roles, and a function-aware pruning strategy is applied. A shared neuron group is retained to preserve neurons contributing to multiple contexts. Additionally, block-wise sparsity is allocated according to functional complexity, with more complex blocks assigned lower sparsity.}
\label{fig:overview}
\end{figure*}

Fig. \ref{fig:overview} illustrates the overall framework of our proposed method. The main contributions of this work are summarized as follows:

\begin{itemize}

\item We analyze the pruning process and identify the failure to distinguish among neurons based on the types of contextual information they process as a key factor limiting generalization.

\item We propose Function-Aware Neuron Grouping (FANG), a post-training structured pruning method that integrates function-aware pruning strategy, shared neuron group retention mechanism, and adaptive sparsity allocation. This is the first method to explicitly consider functional specialization during pruning to enhance generalization.

\item Extensive experiments demonstrate that our method can be effectively combined with classical pruning approaches such as FLAP and OBC, leading to 1.5\%–8.5\% improvements in downstream accuracy while maintaining low perplexity, thereby achieving state-of-the-art (SOTA) performance.

\end{itemize}

\section{Related Works}

\subsection{Structured Pruning}

Pruning refers to the removal of redundant parameters in a model to reduce its storage and computational costs. Based on the granularity of the pruned units, pruning methods can be broadly categorized into unstructured pruning and structured pruning.

Unstructured pruning removes individual weights \cite{obc,sparsegpt,wanda,owl}, resulting in sparse weight matrices. While it often preserves model performance better than structured pruning at the same sparsity level, its practical speedup typically relies on specialized hardware accelerators. In contrast, structured pruning removes entire rows, columns, or even full layers of the weight matrix \cite{flap,ziplm,men2024shortgpt,slicegpt}. This leads to reduced channel or layer counts, enabling actual inference acceleration on general GPUs. 

This work focuses on structured pruning, with the goal of mitigating performance degradation on downstream tasks after pruning.

\subsection{Post-training Pruning without Fine-tuning}

Structured pruning for LLMs is typically implemented in two ways: training-based pruning and training-free pruning.

Training-based pruning can achieve higher compression but requires costly retraining, resulting in substantial computational and storage overhead during this stage \cite{darwinlm,shearedllama}. In contrast, training-free pruning—typically implemented in a post-training manner—is more efficient. It typically adopts a layer-wise pruning strategy, where pruning is performed independently at each layer: unimportant parameters are identified and removed based on predefined importance metrics, followed by numerical compensation to minimize the reconstruction error of that layer's output \cite{ziplm,slicegpt,slimgpt}. This enables the model to recover without further retraining and requires only a small calibration set. Methods such as FLAP \cite{flap}, ModeGPT \cite{modegpt}, and SoBP \cite{sobp} follow this paradigm.

Our study focuses on the post-training pruning paradigm without fine-tuning. Although this approach only requires a small calibration set,  the scarcity of data may give rise to overfitting, which frequently leads to the degraded generalization ability. This has become a critical challenge that this research endeavors to tackle.

\section{Preliminary}

This section introduces the general framework of post-training structured pruning for LLMs, including pruning strategies for Transformer architectures and layer-wise pruning methods.

\subsection{Pruning of Transformer}

Each Transformer \cite{transformer} block consists of two components: a multi-head attention (MHA) module and a feed-forward network (FFN). In each layer $l$, MHA pruning is typically performed at the head level by removing the entire projection matrices $W^{l,i}_{q}$, $W^{l,i}_{k}$ and $W^{l,i}_{v} \in \mathbb{R}^{d_H \times d} $ for the pruned head $i$, along with the corresponding columns in $W^{l}_{o} \in \mathbb{R}^{d \times d} $. $d$ denotes the embedding dimension, $d_H$ is the hidden dimension of each attention head. Given a pruning mask $M^{l}_{H} \in {\{0, 1\}}^{N_h}$, where 1 indicates a pruned head and 0 indicates a retained head, and $N_h$ denotes the total number of attention heads in a layer. The output of the pruned MHA module is formulated in Eq. \ref{eq:head} and \ref{eq:mha}. Attn refers to the self-attention operation, and $\circ$ represents element-wise multiplication.
\begin{equation}
  X_{\text {head}}^{l, i}=(1 - M_{H}^{l, i}) \circ \operatorname{Attn}\left(X^{l-1}, W_{q,k,v}^{l, i}\right),
  \label{eq:head}
\end{equation}
\begin{equation}
 {X}_{\text {MHA}}^{l}=W_{o}^{l} \operatorname{Concat}\left(X_{\text {head }}^{l, 1}, \ldots, X_{\text {head }}^{l, N_{h}}\right).
  \label{eq:mha}
\end{equation}

In the FFN module, pruning is usually conducted at the neuron level by removing the corresponding rows in $W_{up} \in \mathbb{R}^{d \times N_n} $ and columns in $W_{down} \in \mathbb{R}^{N_n \times d} $. $N_n$ denotes the number of neurons in a layer, which is also the dimension of the hidden features in the FFN. Given a pruning mask $M^{l}_{N} \in {\{0, 1\}}^{N_n}$, the output of the pruned MHA module is formulated in Eq. \ref{eq:ffn}, where $\sigma$ denotes activation function.
\begin{equation}
\begin{aligned}
 {X}_{\text {FFN}}^{l}&=W_{down}^{l} (1 - M_{N}^{l}) \circ \\
 &\sigma\left( W_{up}^{l} ({X}_{\text {MHA}}^{l}+X^{l-1})\right).
  \end{aligned}
  \label{eq:ffn}
\end{equation}

\subsection{Layer-wise Pruning}

Layer-wise pruning is a kind of classical method in post-training pruning. Given a linear layer $l$ with weight matrix $W^{l} \in \mathbb{R}^{C_{out} \times C_{in}} $ and input features $X^{l} \in \mathbb{R}^{C_{in} \times L}$, the goal of layer-wise pruning is to obtain a sparse weight matrix $\widehat{W}$ that minimizes the reconstruction error of the output, as defined in the following equation: 
\begin{equation}
\begin{small}
 \arg \min _{{\widehat{W}}^{l}}\|W^{l}X^{l}-{\widehat{W}}^{l}X^{l}\|_{2}^{2} \quad \text { s.t } \quad sp({\widehat{W}}^{l}) \geq {sp}^{l},
\end{small}
  \label{eq:layerwise}
\end{equation}
where $sp({\widehat{W}}^{l})$ denotes the sparsity of the pruned weights and $sp$ is the target sparsity. This formulation is typically applied to the $W_o$ and $W_{down}$ layers to identify removable input channels, while in the $W_{qkv}$ and $W_{up}$ layers, the corresponding output channels are directly removed according to the pruning decisions made in $W_o$ and $W_{down}$.

A common solution is to define an importance metric for each input channel, remove those with the lowest importance, and apply numerical compensation to reduce reconstruction error. Structured OBC \cite{ziplm} and FLAP \cite{flap} are representative methods of this approach.

\section{Method}

The proposed method adopts a function-aware pruning strategy by grouping neurons according to their functional roles and pruning them independently. It enhances robustness by preserving a shared neuron group and adaptively allocating layer-wise sparsity based on functional complexity.

\subsection{Functional Grouping of Neurons Based on Context Clustering}

The proposed method begins by identifying functional neuron groups, which serve as the basis for subsequent pruning. Following prior work, we define functional roles as the model's capacity to process distinct types of semantic context \cite{dejavu,llamamoe}. This gives rise to two key challenges: (1) how to identify different context types from the model’s learned representations, and (2) how to determine which neurons primarily contribute to each type. 

To address these challenges, we propose a clustering-based strategy to distinguish semantic contexts and a score-based assignment mechanism to associate neurons with their corresponding functional groups.

\subsubsection{Context Clustering.} We cluster input tokens to distinguish different types of contextual information. Specifically, we extract the input to each FFN layer, defined as ${X}_{\text {MHA}}^{l}+X^{l-1}$ in Eq. \ref{eq:ffn}, and apply the K-Means algorithm to partition them into K clusters $\{C_{1}^{l}, C_{2}^{l}, \dots, C_{K}^{l}\}$. To reduce computational complexity, we first apply Principal Component Analysis (PCA) to reduce the dimensionality of the token representations, preserving the most informative components for clustering. 

\subsubsection{Cluster–Neuron Score Matrix $S$.} Once token clusters are obtained, we quantify the contribution of each neuron to each cluster using a Taylor expansion-based sensitivity score \cite{oneordertaylor,importancetaylor,obd}. For the $k$-th token cluster and the $j$-th neuron, the score is computed as:
\begin{equation}
 s_{k,j}^{l} = \frac{1}{N_{C_{k}}} \sum_{i \in C_{k}}\left|h_{i, j}^{l} \frac{\partial \mathcal{L}}{\partial h_{i, j}^{l}}\right|,
  \label{eq:score}
\end{equation}
where $i \in C_{k}$ denotes the tokens belonging to the $k$-the cluster, and $N_{C_{k}}$ is the number of such tokens. $\mathcal{L}$ is the loss function of the pretraining task, and $h^{l}$ denotes the intermediate hidden features in the FFN layer $l$. The resulting score matrix $S^{l} \in \mathbb{R}^{K \times N_n}$ captures the relevance of each of the $N_n$ neurons to each token cluster.

\subsubsection{Neuron-to-Function Assignment using $S$.} We formulate the assignment of neurons to functional groups as a Linear Assignment Problem (LAP). Following the solution to LAP as in \cite{llamamoe}, K neuron groups $\{G_{1}^{l}, G_{2}^{l}, \dots, G_{K}^{l}\}$ are obtained for layer $l$, where each group $G_{k}^{l}$ consists of an equal number of  neurons most relevant to cluster $C_{k}^{l}$. These groups are treated as functionally specialized units, each responsible for processing a specific type of context.

\subsection{Function-Aware Pruning Strategy}

The core idea of the function-aware pruning strategy is to prune neurons independently within each functional neuron group and to compute group-specific importance metrics that place greater emphasis on the tokens each group is responsible for processing.

\subsubsection{Group-Wise Pruning.} Given the neuron groups $\{G_{1}^{l}, G_{2}^{l}, \dots, G_{K}^{l}\}$ , we partition the weight matrix $W_{down}^{l}$ of each FFN layer accordingly. Each sub-matrix $W_{:,G_k}^{l}$ is then pruned independently using importance metrics specific to group $G_{k}^{l}$.

\subsubsection{Token-Aware Neuron Importance Reweighting.} To ensure that each group's pruning process focuses on relevant context, we introduce a reweighting mechanism during importance estimation. Token clusters that are semantically closer to the cluster associated with a neuron group are assigned higher weights $\alpha$. To quantify the semantic relevance between token clusters, we compute the pairwise $L_2$ distances $d_{k,j}^{l}$ between cluster centers, which are measured at the input to $W_{down}^{l}$, resulting in a distance matrix $D^{l}\in \mathbb{R}^{K \times K}$. For each token cluster $k$, the relevance weights ${\alpha}_k^{l} \in \mathbb{R}^{K}$ are computed via a softmax over the negative distances:
\begin{equation}
\alpha_k^{l} = \text{softmax}\left(-\frac{d_k^{l}}{\tau}\right),
\label{eq:softmax}
\end{equation}
where $\tau$ is a temperature hyperparameter that controls the sharpness of the distribution.

\subsubsection{Group-Wise Reweighted Pruning Objective.} Incorporating group-wise structure and semantic relevance, the standard layer-wise pruning objective (Eq. \ref{eq:layerwise}) is reformulated as:

\begin{equation}
\begin{small}
\begin{aligned}
\underset{{\widehat{W}}^{l}}{\arg\min} \sum_{k,j}& \alpha_{k,j}^{l} 
\left\| W_{:,G_k}^{l} X_{G_k,C_j}^{l} - \widehat{W}_{:,G_k}^{l} X_{G_k,C_j}^{l} \right\|_2^2 \\
&\text{s.t. } sp(\widehat{W}_{:,G_k}^{l}) \geq {sp}_{G_k^{l}}, \forall G_k^{l},
\end{aligned}
\end{small}
\label{eq:groupprune}
\end{equation}

where $W_{:,G_k}$ refers to the columns of the weight matrix corresponding to neurons in $G_k$, and $X_{G_k,C_j}$ refers to the sub-matrix of input features indexed by neurons in $G_k$ and tokens in $C_j$. To address the above objective, OBC \cite{ziplm} or FLAP \cite{flap} can be applied independently to each group. The neuron importance and compensation within a group are computed with consideration of $\alpha_{k,j}^{l}$.

\begin{table*}[htbp]
\small
\centering
\begin{tabular}{c|l|cc|cc|cc|cc}
\toprule
\multicolumn{2}{c|}{\textbf{Model}}
& \multicolumn{2}{c|}{\textbf{LLaMA1-7B}} 
& \multicolumn{2}{c|}{\textbf{LLaMA2-7B}} 
& \multicolumn{2}{c|}{\textbf{LLaMA2-13B}} 
& \multicolumn{2}{c}{\textbf{LLaMA2-70B}} \\
\cmidrule(lr){1-2} \cmidrule(lr){3-4} \cmidrule(lr){5-6} \cmidrule(lr){7-8} \cmidrule(lr){9-10}
 Sparsity & \hspace{5em}Method
 & PPL$\downarrow$ & Avg$\uparrow$ 
 & PPL$\downarrow$ & Avg$\uparrow$ 
 & PPL$\downarrow$ & Avg$\uparrow$ 
 & PPL$\downarrow$ & Avg$\uparrow$ \\
\midrule

\multirow{1}{*}{0\%} 
& Dense & 5.68 & 66.05 
         & 5.47 & 66.69 
         & 4.88 & 69.24 
         & 3.32 & 73.61 \\
         
\midrule

\multirow{7}{*}{ } 
& SliceGPT (ICLR 2024) & 6.99 & 56.16  & 6.84 & 54.25 & 6.06 & 56.78 & 4.46 & 69.60 \\
& SoBP (EMNLP 2024)  & 6.78 & 62.19 & 6.53 & \underline{63.27}& 5.62 & \underline{67.73} & \textbf{3.88} & 71.24 \\
& SVD-LLM (ICLR 2025) & 7.89 & 56.25 & 8.38 & 48.70 & 6.66 & 58.98 & 4.66 & 68.14 \\
% \rowcolor{green!10}
\cellcolor{white}{20\%}& FLAP (AAAI 2024)   & 6.89 & 61.19 & 7.16 & 56.62 & 6.31 & 61.55 & 4.12 & 71.60 \\
% \rowcolor{green!20}
& \textbf{F-FANG (Ours)}   & 6.70 & \textbf{63.21} & 6.63 & \textbf{63.29} & 5.82 & \textbf{67.91} & 3.98 & \underline{72.11} \\
% \rowcolor{blue!10}
& OBC (NeurIPS 2023) &\underline{6.61} &61.53 &\textbf{6.30} &62.27 &\textbf{5.53} &66.59 &\underline{3.95}	&70.54 \\
% \rowcolor{blue!20}
& \textbf{O-FANG (Ours)} &\textbf{6.43} &\underline{62.65} &\underline{6.31}	&62.88&\underline{5.57}	&67.06 &3.97	&\textbf{72.24} \\
% \cellcolor{white}

\midrule

\multirow{7}{*}{ } 
& SliceGPT \cite{slicegpt} & 8.69 & 46.90  & 8.64 & 46.70 & 7.44 & 50.10 & 5.41 & 61.61 \\
& SoBP \cite{sobp}  & 7.57 & \underline{59.61} & 7.58 & 59.15& \underline{6.27} & \textbf{66.82} & \textbf{4.36} & 70.30 \\
& SVD-LLM \cite{svdllm} & 9.52 & 51.39 & 10.66 & 44.77 & 8.00 & 53.16 & 5.44 & 64.65 \\
% \rowcolor{green!10}
{30\%}& FLAP \cite{flap}   & 8.23 & 57.30 & 8.85 & 50.91 & 7.57 & 57.27 & 4.82 & 69.68 \\
% \rowcolor{green!20}
& \textbf{F-FANG (Ours)}   & 7.94  &\textbf{60.34} & 7.78 & \underline{59.44} & 6.53 & \underline{65.50} & 4.46 & \underline{71.13} \\
% \rowcolor{blue!10}
& OBC \cite{ziplm} &\underline{7.38}	&57.56 &\underline{7.34}	&57.88 &7.72	&51.56 &\underline{4.40}	&67.94\\
% \rowcolor{blue!20}
& \textbf{O-FANG (Ours)} &\textbf{7.23}	&59.50 &\textbf{7.23}	&\textbf{59.83} &\textbf{6.17}	&64.21 &4.46	&\textbf{71.34}\\

\midrule

\multirow{6}{*}{ } 
& SliceGPT \cite{slicegpt} & 15.94 & 39.64  & 12.80 & 41.47 & 10.60 & 44.31 & 7.08 & 52.00 \\
& SoBP \cite{sobp}  & 9.09 & \textbf{56.10} & 9.28 & \textbf{56.06}& \underline{7.39} & \underline{60.86} & \underline{4.96} & 68.58 \\
& SVD-LLM \cite{svdllm} & 13.85 & 42.95 & 16.14 & 40.47 & 10.79 & 45.40 & 6.83 & 58.05 \\
% \rowcolor{green!10}
{40\%}& FLAP \cite{flap}   & 10.16 & 52.45 & 11.49 & 48.70 & 9.07 & 53.18 & 6.24 & 67.96 \\
% \rowcolor{green!20}
& \textbf{F-FANG (Ours)} & 10.31 & 53.82 & 10.34 & 50.94 & 7.84 & 58.62 & 5.06 & \textbf{69.30} \\
% \rowcolor{blue!10}
& OBC \cite{ziplm} &\underline{8.85}	&52.49 &\underline{9.13}	&52.47 &11.67	&48.80 &\textbf{4.87}	&67.23 \\
% \rowcolor{blue!20}
& \textbf{O-FANG (Ours)} &\textbf{8.56} &\underline{55.16} &\textbf{8.67}	&\underline{54.96} &\textbf{7.04}	&\textbf{61.16} &5.01	&\underline{68.77} \\

\bottomrule
\end{tabular}
\caption{Comparison of Structured Pruning and Low-Rank Decomposition Methods on LLaMA Models. \textbf{Bold} indicates the best result, \underline{underline} denotes the second-best. O-FANG and F-FANG denote the combinations of our method with OBC and FLAP, respectively.}
\label{tab:llama-compression}
\end{table*}

\subsection{Enhancing Robustness and Performance}

To enhance the robustness and performance of the function-aware pruning strategy, two limitations must be addressed. First, some neurons contribute to multiple context types and should not be rigidly assigned to a single functional group; pruning such shared neurons may lead to the loss of general-purpose capacity. Second, functional neuron groups vary in complexity, and applying a uniform sparsity ratio may cause unnecessary accuracy loss in more critical groups.

To mitigate these issues, we introduce a shared neuron group retention mechanism and an adaptive sparsity allocation strategy that assigns block-wise sparsity based on the aggregated functional complexity within each block.

\subsubsection{Shared Neuron Group Retention.} To preserve general-purpose capacity, we construct one shared neuron group per layer using the cluster–neuron score matrix $S$. The group size equals that of each functional group, denoted as $m = N_n / (K + 1)$. For each token cluster, we select the top-$m$ scoring neurons. Neurons selected by multiple clusters are ranked by selection frequency, and the top-$m$ ones are retained to form the shared group. These neurons are exempted from pruning, ensuring that widely useful representations are preserved.

\subsubsection{Adaptive Sparsity Allocation.} Inspired by ShortGPT \cite{men2024shortgpt}, we define the Functional Complexity (FC) of each block as the degree of change between its input and output representations, measured by cosine similarity. A lower similarity indicates higher complexity, suggesting that the block performs more substantial transformations. Formally, the FC of block $l$ is computed as:
\begin{equation}
{FC}^{l} = 1 - \mathbb{E}_{X, t} \frac{X_{:, t}^{l \top} X_{:, t}^{l+1}}{\left\|X_{:, t}^{l}\right\|_{2}\left\|X_{:, t}^{l+1}\right\|_{2}},
\label{eq:fc}
\end{equation}

where $X_{:, t}^{l}$ denotes the representation of token $t$ at the input of block $l$. Following OWL \cite{owl}, we set the block-wise sparsity $sp^{l} \propto 1 - {FC}^{l}$, and linearly scale all $sp^{l}$ values to fall within $[0.5sp, 1.5sp]$, where $sp$ denotes the target sparsity.

\begin{table*}[htbp]
\small
\centering
\begin{tabular}{c|l|c|c|c|c|c|c|c}
\toprule
\textbf{Sparsity} & \hspace{6em}\textbf{Method} &\textbf{BoolQ} &\textbf{HellaS.}	&\textbf{WinoG.}	&\textbf{ARC-e}	&\textbf{ARC-c}	&\textbf{OBQA}	&\textbf{PIQA}\\
\midrule

\multirow{1}{*}{0\%} 
& Dense & 77.74 & 76.02 & 68.98 & 74.58 & 46.25 & 44.20 & 79.05  \\
         
\midrule

\multirow{7}{*}{30\%} 
& FLAP (AAAI 2024)  & 44.71 & 56.58 & 61.72 & 53.83 & 31.23 & 37.00 & 71.27 \\
& SliceGPT (ICLR 2024) & 38.32 & 49.09  & 60.69 & 49.58 & 31.74 & 32.60 & 64.91 \\
& SoBP (EMNLP 2024)  & \textbf{71.19} & 67.27 & 66.22 & 59.81& 37.63 & 38.40 & 73.50 \\
& LLM surgeon (ICLR 2024) & - & 60.72 & 61.09 & 63.09 & 36.69 & - & \textbf{73.56}\\
& ModeGPT (ICLR 2025) & - & 63.26 & 67.32 & 63.26 & 38.73 & - & 70.40\\
% \rowcolor{blue!10}
& OBC (NeurIPS 2023) &64.68 &64.34	&66.06	&62.88	&36.43	&38.20	&72.58 \\
% \rowcolor{blue!20}
& \textbf{O-FANG (Ours)} &63.06	&\textbf{67.33}	&\textbf{67.32}	&\textbf{67.34}	&\textbf{39.93}	&\textbf{40.40}	&73.45\\

\midrule

\multirow{5}{*}{40\%}
& SliceGPT \cite{slicegpt} & - & 34.80 & 53.43 & 36.49 & 24.57 & - & 54.90 \\
& LLM surgeon \cite{Surgeon}  & - & 48.04 & 54.38 & 52.31 & 30.29 & - & \textbf{69.26}\\
& ModeGPT \cite{modegpt} & - & 53.01 & 61.96 & 49.45 & 30.03 & - & 64.96\\
% \rowcolor{blue!10}
& OBC \cite{ziplm} &\textbf{63.00} &52.04	&60.62	&55.68	&31.14	&37.60	&67.19\\
% \rowcolor{blue!20}
& \textbf{O-FANG (Ours)} &61.19	&\textbf{58.72}	&\textbf{64.48}	&\textbf{58.16}	&\textbf{33.11}	&\textbf{40.20}	&68.82\\

\bottomrule
\end{tabular}
\caption{Zero-Shot Accuracy of Compressed LLaMA2-7B on Downstream Tasks.}
\label{tab:downstream}
\end{table*}

\section{Experiments}

\subsection{Experimental Settings}

\subsubsection{Implementation Details.} In the proposed method, neuron functional grouping, the Function-Aware Pruning Strategy, and shared neuron group retention are applied only to FFN pruning, while adaptive sparsity allocation is used for both attention head and FFN pruning. Each layer is divided into seven functional groups and one shared group. For context clustering, input token representations are reduced to 64 dimensions via PCA, where the chosen dimensionality balances clustering efficiency and accuracy. In Function-Aware Pruning, $\tau$ controls reweighting, with model-specific values (e.g., $\tau=9$ for LLaMA2-7B and $\tau=7$ for LLaMA3.1-8B across all sparsity levels). We explore $\tau$ values in the range of 3 to 11 with a step size of 2 to select optimal values. The calibration set is sampled from WikiText2 \cite{wikitext} training data, which is the most commonly used calibration dataset in the model compression literature \cite{flap,svdllm,sobp}. All steps use 128 samples with a sequence length of 2048.

\subsubsection{Evaluation.} We conduct experiments on LLaMA-1 \cite{llama}, LLaMA-2 \cite{llama2}, LLaMA-3.1 \cite{llama3}, and Qwen-2.5 \cite{qwen}, covering model sizes from 7B to 70B and sparsity levels from 20\% to 40\%. Perplexity is evaluated on the WikiText2 \cite{wikitext} test set. Downstream performance is measured by average accuracy across standard benchmarks, including ARC-c, ARC-e \cite{arc}, WinoGrande \cite{winogrande}, BoolQ \cite{boolq}, HellaSwag \cite{hellaswag}, OpenBookQA \cite{openbook}, and PIQA \cite{piqa}. These tasks follow standard evaluation practices in model compression. Additionally, we use the MMLU \cite{mmlu} benchmark to more comprehensively assess model capabilities.

\subsection{Main Results}
\subsubsection{Perplexity and Zero-Shot Accuracy.} Tab.~\ref{tab:llama-compression} compares different methods on LLaMA models in terms of perplexity and average zero-shot accuracy across 7 downstream tasks. Our method exhibits strong adaptability and can be effectively integrated with advanced pruning strategies such as OBC and FLAP, resulting in O-FANG and F-FANG that further enhance overall performance. It consistently achieves lower perplexity while improving accuracy, performing comparably to or better than accuracy-leading methods such as SoBP. In most settings, perplexity remains lower, with only minor increases in a few cases. Overall, it delivers the best overall performance, consistently ranking among the top across both metrics.

Tab. \ref{tab:downstream} reports the zero-shot accuracy on downstream tasks for the compressed LLaMA2-7B model using different pruning methods. O-FANG improves accuracy by 1\%–5\% over OBC on nearly all tasks, with only a slight drop on BoolQ. It also outperforms other advanced methods such as LLM surgeon and ModeGPT. These results demonstrate the robustness and generalization advantages of the proposed approach.

\subsubsection{Results on More Advanced LLMs and benchmark.} We further evaluate our method on Qwen-2.5 and LLaMA-3.1 models. Results show that our approach consistently improves the downstream accuracy of OBC by a significant margin, even on these more advanced LLMs. This demonstrates the broad applicability of the proposed method.

The MMLU \cite{mmlu} dataset is designed to evaluate the capabilities of LLMs across a wide range of language understanding tasks. It serves as a robust benchmark for assessing the generalization ability. As shown in Tab \ref{tab:mmlu}, integrating our method improves the baseline's zero-shot accuracy on MMLU, indicating enhanced generalization performance.

\begin{table}[htbp]
\small
\centering
\begin{tabular}{c|c|l|cc}
\toprule
\textbf{Model} &\textbf{Sparsity} & \hspace{0.2em}\textbf{Method} &\textbf{PPL}$\downarrow$ &\textbf{Avg}$\uparrow$\\
\midrule

\multirow{2}{*}{\makecell[c]{LLaMA-3.1 \\ 8B}} 
& \multirow{2}{*}{20\%} & OBC & 9.17 & 56.88\\
&  & O-FANG  &\textbf{8.31}  &\textbf{64.53}\\

\midrule

\multirow{2}{*}{\makecell[c]{Qwen2.5 \\ 7B}} 
& \multirow{2}{*}{40\%} & OBC & 14.04 & 48.82\\
&  & O-FANG  &\textbf{11.72}  &\textbf{52.14}\\

\bottomrule
\end{tabular}
\caption{Performance of Pruned LLaMA-3.1-8B and Qwen2.5-7B models.}
\label{tab:qwenllam3}
\end{table}

\begin{table}[htbp]
\small
\centering
\begin{tabular}{c|c|l|c}
\toprule
\textbf{Model} &\textbf{Sparsity} & \hspace{0.2em}\textbf{Method} & \textbf{Acc}$\uparrow$\\
\midrule

\multirow{2}{*}{\makecell[c]{LLaMA2-7B}} 
& \multirow{2}{*}{30\%} & OBC & 24.68\\
&  & O-FANG &\textbf{26.25}\\

\midrule

\multirow{2}{*}{\makecell[c]{LLaMA2-13B}} 
& \multirow{2}{*}{30\%} & OBC & 38.08\\
&  & O-FANG &\textbf{42.37}\\

\bottomrule
\end{tabular}
\caption{Zero-shot Accuracies of compressed models on the MMLU benchmark.}
\label{tab:mmlu}
\end{table}

\subsection{Ablation Experiments}

\subsubsection{Ablation Study on Key Components.} The ablation results in Tab. \ref{tab:ablation} highlight the contribution of each module to overall performance. Experiments are conducted on the LLaMA2-7B model. Adaptive Sparsity Allocation (ASA) achieves a better balance between accuracy and sparsity, improving accuracy by approximately 1.3\% over the baseline. Incorporating Shared Neuron Group Retention (SNGR) further improves accuracy by around 0.5\%, as it preserves neurons that are important for handling diverse context types. Applying Function-Aware Pruning (FAP) provides an additional gain of 0.2\%-0.6\%, reflecting improved generalization. The full method, combining all modules, achieves the highest accuracy at both sparsity levels, confirming their complementary benefits.

\begin{table}[htbp]
\small
\centering
\begin{tabular}{ccc|cc}
\toprule
\textbf{ASA} &\textbf{SNGR} & \textbf{FAP} &\textbf{30\%} &\textbf{40\%}\\
\midrule
&&& 57.88 & 52.47\\
\checkmark & & & 59.17 (+1.29)& 53.90(+1.43)\\
\checkmark &\checkmark & &59.64 (+0.47)& 54.38 (+0.48)\\
\checkmark &\checkmark &\checkmark &\textbf{59.83} (+0.19)& \textbf{54.96} (+0.58)\\

\bottomrule
\end{tabular}
\caption{Ablation study on LLaMA2-7B under 30\% and 40\% sparsity. The three abbreviations refer to Adaptive Sparsity Allocation (ASA), Shared Neuron Group Retention (SNGR), and Function-Aware Pruning (FAP). Average zero-shot average accuracy of 7 downstream tasks is reported.}
\label{tab:ablation}
\end{table}

\subsubsection{Grouping and Re-weighting Method.} Tab. \ref{tab:fap} compares our proposed neuron grouping method with a random grouping baseline, where neurons in each layer are randomly assigned to groups and one group is randomly selected as the shared group. The results show that our method more effectively clusters neurons with similar functions, enabling the function-aware pruning strategy to work as intended.

Tab. \ref{tab:fap} compares several reweighting strategies. Reverse assigns higher weights to less semantically related token clusters; Uniform uses equal weights; Only-Matched uses only the assigned token cluster for each group. The average accuracy follows: Ours $>$ Uniform $>$ Reverse $>$ Only-Matched. These results demonstrate that the proposed reweighting strategy effectively helps each group focus on neurons most relevant to its function. The low accuracy of Only-Matched is due to limited token coverage during pruning.

\begin{table}[htbp]
\small
\centering
\begin{tabular}{c|c|cc}
\toprule
\textbf{Experiment} &\textbf{Method} &\textbf{30\%} &\textbf{40\%}\\
\midrule
\multirow{2}{*}{\makecell[c]{Grouping}} 
&  Random & 59.20 & 53.94\\
&  Ours  &\textbf{59.83}  &\textbf{54.96}\\
\midrule
\multirow{4}{*}{\makecell[c]{Re-weight}} 
&  Reverse & 59.05 & 52.44\\
&  Uniform  &59.47  &54.77\\
&  Only-Matched  &55.90  & 51.01\\
&  Ours  &\textbf{59.83}  &\textbf{54.96}\\
\bottomrule
\end{tabular}
\caption{Ablation of Function-Aware Pruning Strategy: Grouping and Reweighting Methods. Average zero-shot average accuracy of 7 downstream tasks is reported and tested on LLaMA2-7B.}
\label{tab:fap}
\end{table}

\subsubsection{Adaptive Sparsity Allocation.} Different strategies for allocating sparsity across blocks are compared in Tab. \ref{tab:asa}. We evaluate a Taylor-based method, which assigns sparsity based on output sensitivity, and our approach, which uses functional complexity. Experimental results show that our approach yields higher average accuracy, suggesting that functional complexity is a more effective criterion for sparsity allocation than output sensitivity.

\begin{table}[htbp]
\small
\centering
\begin{tabular}{c|c|cc}
\toprule
\textbf{Model} &\textbf{Sparsity} &\textbf{Taylor} &\textbf{FC (Ours)}\\
\midrule
 
LLaMA2-7B& 30\% & 58.85 & \textbf{59.83}\\
LLaMA2-13B& 40\% & 52.17& \textbf{61.16}\\

\bottomrule
\end{tabular}
\caption{Effectiveness of Functional Complexity in Adaptive Sparsity Allocation. Average zero-shot average accuracy of 7 downstream tasks is reported.}
\label{tab:asa}
\end{table}

\subsection{Additional Experimental Analysis} 

\subsubsection{Compression Time.} 
\begin{figure}[t]
\centering
\includegraphics[width=0.34\textwidth]{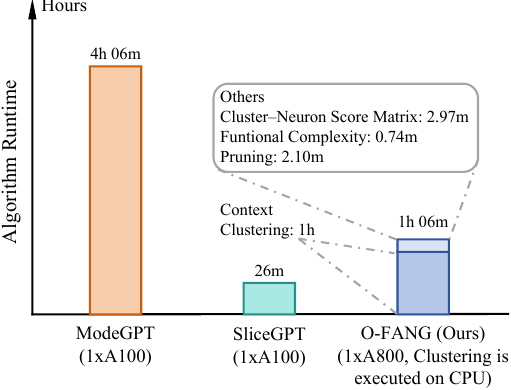} % Reduce the figure size so that it is slightly narrower than the column. Don't use precise values for figure width.This setup will avoid overfull boxes.
\caption{Execution Efficiency Comparison of Different Algorithms (LLaMA2-7B, 40\% Sparsity).}
\label{fig:time}
\end{figure}

Fig.\ref{fig:time} presents the execution time analysis of algorithms on LLaMA2-7B. In our method, context clustering is the most time-consuming step, mitigated by PCA-based dimensionality reduction, which keeps the total time within one hour. Although less efficient than the fastest baseline, the runtime remains substantially lower than ModeGPT, indicating acceptable time cost. As shown in Tab.\ref{tab:downstream}, our method achieves higher accuracy than both baselines under this acceptable cost.

\subsubsection{Comparison with Downstream-Data-Based Calibration.} Another way to improve downstream accuracy is to use downstream task data as the calibration set. Tab. \ref{tab:calibration} compares this approach with ours. The main difference is the calibration source: the baseline uses Alpaca (instruction-tuning data) \cite{alpaca}, while our method uses WikiText (pretraining data) with function-aware pruning. All other settings, including sparsity allocation and the number of calibration tokens, are kept consistent. Results show that both approaches achieve improvements in downstream accuracy, but our method is more effective at preserving perplexity.

\begin{table}[htbp]
\small
\centering
\begin{tabular}{c|c|c|cc}
\toprule
\textbf{Settings} &\textbf{Method} &\textbf{Calibration}&\textbf{PPL} $\downarrow$  &\textbf{Avg} $\uparrow$ \\
\midrule
\multirow{2}{*}{\makecell[c]{7B\\30\%}} 
&  OBC+ASA & Alpaca &10.24 & 60.35\\
&  O-FANG & WikiText &7.23  &59.83\\
\midrule
\multirow{2}{*}{\makecell[c]{13B\\30\%}} 
&  OBC+ASA & Alpaca &9.35 & 64.30\\
&  O-FANG & WikiText &6.17 &64.21\\
\bottomrule
\end{tabular}
\caption{Downstream Calibration vs. Function-Aware Pruning. Results are reported for OBC with Adaptive Sparsity Allocation (ASA) and O-FANG, evaluated on LLaMA2.}
\label{tab:calibration}
\end{table}

\section{Conclusion}

In this work, we propose Function-Aware Neuron Grouping (FANG), a post-training pruning framework inspired by the functional specialization observed in LLMs. By grouping neurons based on functional roles, reweighting importance scores, retaining shared neurons, and adaptively allocating sparsity, FANG improves generalization without compromising efficiency. Experiments across multiple models and sparsity levels show that FANG outperforms existing methods such as FLAP and OBC, achieving 1.5\%-8.5\% higher downstream accuracy while maintaining low perplexity.

\section{Acknowledgements}

This work was supported by the National Key R\&D Program of China under Grant No. 2022ZD0160601, and National Natural Science Foundation of China (Grant No. 62276260, 62206290).

\bibliography{main}

\newpage

% \ifreproStandalone
% \setlength{\pdfpagewidth}{8.5in}
% \setlength{\pdfpageheight}{11in}
% % \usepackage{algorithm}
% % \usepackage{algpseudocode}
% \urlstyle{rm} % DO NOT CHANGE THIS
% \def\UrlFont{\rm}  % DO NOT CHANGE THIS
% \fi
% \makeatletter
% \@ifundefined{isappendixMainFile}{
%   % We are compiling a standalone document
%   \newif\ifreproStandalone
%   \reproStandalonetrue
% }{
%   % We are being \input into the main paper
%   \newif\ifreproStandalone
%   \reproStandalonefalse
% }
% \makeatother

% \frenchspacing  % DO NOT CHANGE THIS

% \begin{document}

\section{Appendix}

\begin{table*}[htbp]
\small
\centering
\setlength{\tabcolsep}{3pt}
\begin{tabular}{l|cc|cc|cc}
\toprule
\hspace{3.5em}\textbf{Model and Sparsity}
& \multicolumn{2}{c|}{\textbf{LLaMA2-7B 30\%}} 
& \multicolumn{2}{c|}{\textbf{LLaMA2-7B 40\%}} 
& \multicolumn{2}{c}{\textbf{LLaMA2-13B 40\%}} \\
\midrule
\hspace{6.5em}Method
 & PPL$\downarrow$ & Avg$\uparrow$ 
 & PPL$\downarrow$ & Avg$\uparrow$ 
 & PPL$\downarrow$ & Avg$\uparrow$ \\
\midrule

{Traditional OBC} & 7.23 &58.10 &9.20 &51.45 &9.98 &47.48 \\
{Variant of OBC (Ours)} &7.34 &57.88 &9.13 &52.47 &11.67 &48.82 \\
{Traditional OBC + Sparsity Allocation} &7.05 &58.81 &8.31 &53.40 &6.85 &58.63\\
{Variant of OBC + Sparsity Allocation} &7.12 &59.31 &8.37 &53.90 &6.86 &59.37 \\

\bottomrule
\end{tabular}
\caption{Comparison of Traditional OBC and Our Variant Under Different Sparsity Settings and Sparsity Allocation Strategies.}
\label{tab:obcvariant}
\vspace{-6mm}
\end{table*}

\subsection{Extending OBC with Function-Aware Pruning}

This section explains how the function-aware pruning strategy is applied to OBC. We begin by introducing structured OBC pruning, followed by a variant of OBC that we propose. This variant is more efficient than the standard OBC while also yielding improved accuracy on downstream tasks. All OBC-related experiments in this paper are conducted based on this variant. Finally, we describe how neuron importance and compensation values are computed when solving the Group-Wise Reweighted Pruning Objective.

\subsubsection{The Variant of OBC.} For the layer-wise pruning problem, the traditional OBC adopts a greedy iterative pruning strategy. At each step, the importance of all neurons is evaluated, and the least important neuron is pruned and compensated. The matrix $H^{-1}$ is then updated, and neuron importance and compensation are recalculated in the next step. This process is repeated until the target sparsity is reached, as illustrated in Algorithm 1.

\begin{algorithm}[htbp]
\small
\caption{Traditional OBC}
\KwIn{Weight matrix $W$, input matrix $X$, target sparsity $sp$}
\KwOut{Pruned and compensated weight matrix $\widehat{W}$}

Compute Hessian $H = XX^{\top}$\;

Initialize pruning mask $M = \mathbf{0}$\;

\While{number of pruned neurons $< sp$}{
    \For{each input channel $j$}{
        Compute importance $e_j = \sum_{i} \frac{W_{i,j}^2}{H_{j,j}^{-1}}$\;
    }
    Find index $j^\ast = \arg\min_j e_j$\;
    
    Set $M_{j^\ast} = 1$ and prune channel $j^\ast$\;
    
    Update $W$: $W \leftarrow W - \frac{W_{:,j}}{H_{j,j}^{-1}} H_{j,:}^{-1}$\;

    Update $H^{-1} \leftarrow H^{-1} $\;
}
\Return{$\widehat{W}=W$}
\label{alg:obc}
\end{algorithm}

Instead of performing iterative pruning and compensation, our proposed OBC variant removes all unimportant neurons in a single step and applies a unified compensation. Specifically, neuron importance is calculated using Eq.\ref{eq:error}, and the neurons with the lowest importance scores are selected for removal based on the target sparsity. A binary pruning mask $M$ is then constructed, where entries corresponding to pruned neurons are set to 1 and the remaining entries are set to 0. Numerical compensation is subsequently applied according to Eq.\ref{eq:update}.
\begin{equation}
 e_{j}=\sum_{i=1}^{C_{out}} \frac{W_{i, j}^{2}}{H_{j, j}^{-1}},
  \label{eq:error}
\end{equation}
\begin{equation}
 \delta W=-\left[H^{-1} I_{M}\left[H_{M M}^{-1}\right]^{-1} W_{M}^{\top}\right]^{\top}.
  \label{eq:update}
\end{equation}

$H$ is the Hessian matrix, approximated by $XX^{\top}$. $I_M$ and $W_M$ denote the columns of the identity matrix $I$ and the weight matrix $W$ selected by the mask $M$, respectively. $H_{M M}^{-1}$  is the sub-matrix of $H^{-1}$ corresponding to the rows and columns where $M = 1$. The complete procedure is presented in Algorithm 2.

\begin{algorithm}[htbp]
\small
\caption{Our Variant of OBC}
\KwIn{Weight matrix $W$, input matrix $X$, target sparsity $sp$}
\KwOut{Pruned and compensated weight matrix $\widehat{W}$}

Compute Hessian $H = XX^{\top}$\;

\For{each input channel $j$}{
    Compute importance $e_j = \sum_{i} \frac{W_{i,j}^2}{H_{j,j}^{-1}}$\;
}

Sort $e_j$ and determine pruning mask $M$ to keep top-$1{-}sp$ fraction of channels\;

Prune all channels where $M_j = 1$\;

Compute unified compensation $\delta W$ using Eq. \ref{eq:update}\;

Apply update: $\widehat{W} = W + \delta W$\;

\Return{$\widehat{W}$}
\end{algorithm}

Although our variant improves computational efficiency, it does not cause noticeable performance degradation and can even enhance generalization. As shown in Tab. \ref{tab:obcvariant}, under the proposed sparsity allocation strategy, our variant achieves comparable perplexity to standard OBC while yielding higher average zero-shot accuracy. This result may be attributed to traditional OBC's tendency to overfit. Therefore, considering both efficiency and performance, all OBC-related experiments in this work are conducted using our variant.

\subsubsection{Improving OBC to Solve New Pruning Objective.} To apply the proposed OBC variant to the Group-Wise Reweighted Pruning Objective, the weight matrix is first partitioned into sub-matrices $W_{:,G_k}^{l}$ based on $\{G_{1}^{l}, G_{2}^{l}, \dots, G_{K}^{l}\}$. Each sub-matrix is then pruned independently following the OBC procedure described in Eq. \ref{eq:error} and \ref{eq:update}. The key modification lies in the computation of the Hessian matrix $H_k^{l}$, which is redefined for each group $G_{k}^{l}$, using weighted token features:
\begin{equation}
H_k^{l} = \sum_{j}^{K} \alpha_{k,j}^{l} X_{G_k,C_j}^{l} X_{G_k,C_j}^{l \top}.
\label{eq:reweighthessian}
\end{equation}

\subsection{Extending FLAP with Function-Aware Pruning}

The core idea of FLAP is to measure the importance of neurons and attention heads using a fluctuation-based metric, and to apply compensation by introducing bias terms into the $W_{down}$ and $W_o$ layers, as illustrated in the following formulation.

Specifically, Eq.\ref{eq:mean} computes the mean activation $\overline{X}_{i}^{l}$ for each neuron $j$ across $L$ input tokens in layer $l$. Based on this mean, Eq. \ref{eq:bias} calculates the compensation bias $B_{0}^{l}$, where $M^{l}$ is a binary pruning mask in which a value of 1 indicates a pruned neuron, and $\circ$ denotes element-wise multiplication. Finally, Eq. \ref{eq:fluctuation} defines the importance metrics ${S}_{i}^{l}$ as the product of the activation variance and the squared $L_2$ norm of the corresponding weight vector, capturing the neuron's input fluctuation and weight magnitude.

\begin{equation}
\overline{X}_{i}^{l}=\frac{1}{L} \sum_{j=1}^{L} X_{i, j}^{l},
  \label{eq:mean}
\end{equation}

\begin{equation}
B_{0}^{l}={W}^{\ell}\left({M}^{\ell}\circ \overline{X}^{l}\right),
  \label{eq:bias}
\end{equation}

\begin{equation}
{S}_{i}^{l} \propto \sum_{j=1}^{L}\left({X}_{i,j}^{l}-\overline{X}_{i}^{l}\right)^{2} \cdot\left\|{W}_{:, i}^{l}\right\|_{2}^{2}.
  \label{eq:fluctuation}
\end{equation}

When applying the function-aware pruning strategy, neurons are first grouped according to $\{G_{1}^{l}, G_{2}^{l}, \dots, G_{K}^{l}\}$. For a neuron $i$ belonging to group $G_k$, its importance metrics is modified to a reweighted form, as defined in Eq.\ref{eq:group_fluctuation}. After independently pruning unimportant neurons within each group, the resulting submatrices are combined to reconstruct the pruned weight matrix. A unified compensation bias is then applied using Eq.\ref{eq:bias}.

In Eq.~\ref{eq:group_fluctuation}, $C_m$ denotes the $m$ th token cluster from $\{C_{1}^{l}, C_{2}^{l}, \dots, C_{K}^{l}\}$, and $C_m(j)$ indexes the $j$-th token in that cluster. The weight $\alpha_{k,m}$ reflects the semantic relevance between group $G_k$ and token cluster $C_m$.

\begin{equation}
\small
{S}_{i}^{l} = \sum_{m=1}^{K}\sum_{j=1}^{N(C_m)} \alpha_{k,m}\left({X}_{i,C_m(j)}^{l}-\overline{X}_{i}^{l}\right)^{2} \cdot\left\|{W}_{:, i}^{l}\right\|_{2}^{2}.
  \label{eq:group_fluctuation}
\end{equation}

\subsection{Additional Experimental Results}

To facilitate future research and replication, Tab. \ref{tab:detailresult} provides a more detailed version of the results in the main text, including accuracy across various downstream tasks. The results demonstrate that across most models, sparsity settings, and tasks, the proposed F-FANG and O-FANG consistently achieve higher accuracy than FLAP and OBC.

\begin{table*}[htbp]
\small
\centering
\setlength{\tabcolsep}{3pt}
\begin{tabular}{c|c|l|c|c|c|c|c|c|c|c|c}
\toprule
\textbf{Model} & \textbf{Sparsity}& \textbf{Method} &\textbf{PPL} &\textbf{Avg} &\textbf{BoolQ} &\textbf{HellaS.}	&\textbf{WinoG.}	&\textbf{ARC-e}	&\textbf{ARC-c}	&\textbf{OBQA}	&\textbf{PIQA}\\
\midrule

\multirow{13}{*}{LLaMA1-7B} &
0\% & Dense & 5.68 & 66.05 & 75.02 & 76.22 & 70.01 & 72.90 & 44.62 & 44.40 & 79.16  \\
\cmidrule{2-12}
&\multirow{4}{*}{20\%}
& FLAP   & 6.89 & 61.19 & 71.42	&67.72	&67.46	&67.15	&37.52	&41.98	&75.06\\
&& F-FANG   & 6.70	&63.21	&75.02	&70.96	&68.43	&68.56	&42.15	&41.20	&76.17\\
&& OBC & 6.61	&61.53	&66.57	&70.64	&68.51	&67.05	&40.87	&41.00	&76.06\\
&& O-FANG  & 6.43	&62.65	&75.26	&71.43	&68.19	&65.49	&41.04	&41.20	&75.95 \\
\cmidrule{2-12}
&\multirow{4}{*}{30\%}
& FLAP   & 8.23 & 57.30 & 66.88 & 61.70 & 66.61 & 58.42 & 33.87 & 40.40 & 73.23\\
&& F-FANG   & 7.94	&60.34	&73.12	&65.66	&65.43	&65.66	&37.88	&41.20	&73.39\\
&& OBC & 7.38	&57.56	&63.27	&64.41	&65.59	&60.10	&36.69	&40.60	&72.25\\
&& O-FANG  & 7.23	&59.50	&71.07	&66.09	&67.56	&62.08	&36.26	&40.40	&73.07 \\
\cmidrule{2-12}
&\multirow{4}{*}{40\%}
& FLAP   & 10.16 & 52.45 & 63.46	&54.22	&60.79	&52.25	&30.66	&37.94	&67.84\\
&& F-FANG   & 10.31	&53.82	&65.47	&54.48	&60.77	&57.49	&33.11	&37.00	&68.39\\
&& OBC & 8.85	&52.49	&62.48	&52.55	&61.80	&55.72	&32.42	&36.20	&66.27\\
&& O-FANG  & 8.56	&55.16	&66.27	&57.84	&65.04	&57.11	&33.28	&38.60	&67.95 \\
\midrule

\multirow{13}{*}{LLaMA2-7B} &
0\% & Dense & 5.47 & 66.69 & 77.74 & 76.02 & 68.98 & 74.58 & 46.25 & 44.20 & 79.05  \\
\cmidrule{2-12}
&\multirow{4}{*}{20\%}
& FLAP   & 7.16 & 56.62 & 63.22	&64.78	&64.21	&57.49	&33.40	&40.24	&73.00\\
&& F-FANG   & 6.63	&63.29	&72.66	&71.58	&67.48	&69.87	&42.06	&42.20	&77.15\\
&& OBC & 6.30	&62.27	&67.52	&70.65	&68.82	&69.53	&41.89	&42.20	&75.30\\
&& O-FANG  & 6.31	&62.88	&67.28	&71.79	&71.35	&70.24	&41.47	&41.80	&76.22 \\
\cmidrule{2-12}
&\multirow{4}{*}{30\%}
& FLAP   & 8.85 & 50.91 & 44.71 & 56.58 & 61.72 & 53.83 & 31.23 & 37.00 & 71.27\\
&& F-FANG   &7.88	&59.44	&68.44	&65.19	&64.40	&64.94	&39.76	&39.40	&73.94\\
&& OBC & 7.34	&57.88	&64.68	&64.34	&66.06	&62.88	&36.43	&38.20	&72.58\\
&& O-FANG  & 7.23	&59.83	&63.06	&67.33	&67.32	&67.34	&39.93	&40.40	&73.45 \\
\cmidrule{2-12}
&\multirow{4}{*}{40\%}
& FLAP   & 11.49 & 48.70 & 59.29	&51.01	&57.60	&47.79	&28.32	&33.76	&63.13\\
&& F-FANG   & 10.34	&50.94	&49.6	&55.68	&58.88	&55.68	&31.48	&36.40	&68.82\\
&& OBC & 9.13	&52.47	&63.00	&52.04	&60.62	&55.68	&31.14	&37.60	&67.19\\
&& O-FANG  & 8.67	&54.96	&61.19	&58.72	&64.48	&58.16	&33.11	&40.20	&68.82 \\
\midrule

\multirow{13}{*}{LLaMA2-13B} &
0\% & Dense & 4.88 & 69.24 & 80.58 & 79.37 & 72.30 & 77.48 & 49.23 & 45.20 & 80.52  \\
\cmidrule{2-12}
&\multirow{4}{*}{20\%}
& FLAP   & 6.31 & 61.55 & 67.59	&67.82	&66.89	&69.86	&39.89	&43.04	&75.77\\
&& F-FANG   & 5.82	&67.91	&81.13	&75.95	&72.93	&75.84	&47.10	&44.40	&78.02\\
&& OBC & 5.53	&66.59	&80.98	&74.49	&71.35	&72.52	&45.22	&44.20	&77.37\\
&& O-FANG  & 5.57	&67.06	&79.42	&76.05	&72.22	&73.53	&45.56	&45.00	&77.64 \\
\cmidrule{2-12}
&\multirow{4}{*}{30\%}
& FLAP   & 7.57 & 57.27 & 65.14& 61.86 & 65.59 & 59.43 & 36.26 & 40.20 & 72.42\\
&& F-FANG   & 6.53	&65.50	&79.57	&71.54	&70.01	&44.62	&74.33	&43.00	&75.41\\
&& OBC & 7.72	&51.56	&76.21	&62.70	&64.80	&34.39	&27.47	&37.60	&57.78\\
&& O-FANG  & 6.17	&64.21	&78.32	&72.03	&71.90	&67.68	&41.72	&43.00	&74.81 \\
\cmidrule{2-12}
&\multirow{4}{*}{40\%}
& FLAP   & 9.07 & 53.18 & 63.77	&55.49	&62.08	&52.27	&32.80	&38.24	&67.61\\
&& F-FANG   & 7.84	&58.62	&72.94	&64.04	&64.09	&62.46	&34.22	&40.40	&72.20\\
&& OBC &11.67	&48.80	&67.92	&59.50	&61.25	&32.53	&26.37	&36.60	&57.4\\
&& O-FANG  & 7.04	&61.16	&74.10	&66.28	&69.30	&64.48	&39.08	&42.60	&72.31 \\
\midrule

\multirow{13}{*}{LLaMA2-70B} &
0\% & Dense & 3.32 & 73.61 & 83.70 & 83.81 & 77.98 & 80.98 & 57.25 & 48.80 & 82.75  \\
\cmidrule{2-12}
&\multirow{4}{*}{20\%}
& FLAP   & 4.12 & 71.60 & 81.57	&82.92	&75.13	&78.27	&54.55	&46.72	&82.00\\
&& F-FANG   & 3.98	&72.11	&82.20	&83.90	&75.77	&79.60	&54.27	&46.80	&82.26\\
&& OBC & 3.95	&70.54	&70.20	&83.55	&77.27	&79.10	&54.35	&48.40	&80.90\\
&& O-FANG  & 3.97	&72.24	&83.00	&83.90	&76.32	&79.20	&54.78	&47.60	&80.90 \\
\cmidrule{2-12}
&\multirow{4}{*}{30\%}
& FLAP   & 4.82 & 69.68 &82.78 & 80.41 & 74.66 & 73.86 & 49.32 & 45.80 & 80.90\\
&& F-FANG   & 4.46	&71.13	&82.75	&82.35	&76.16	&78.50	&52.13	&45.60	&80.41\\
&& OBC & 4.40	&67.94	&68.00	&81.65	&75.61	&75.10	&51.79	&45.80	&77.64\\
&& O-FANG  & 4.46	&71.34	&84.20	&82.15	&76.01	&78.30	&53.07	&46.00	&79.65\\
\cmidrule{2-12}
&\multirow{4}{*}{40\%}
& FLAP   & 6.24 & 67.96 & 80.84	&79.04	&72.86	&74.74	&47.52	&42.54	&78.19\\
&& F-FANG   & 5.06	&69.30	&83.75	&80.25	&75.14&	76.45	&47.87	&43.60	&78.07\\
&& OBC & 4.87	&67.23	&67.6	&78.70	&76.80	&74.70	&50.00	&45.80	&76.99\\
&& O-FANG  & 5.01	&68.77	&81.90	&79.50	&75.45	&73.55	&47.87	&45.40	&77.75\\

\bottomrule
\end{tabular}
\caption{Detailed Results of Perplexity and Downstream Task Accuracy.}
\label{tab:detailresult}
\vspace{-3mm}
\end{table*}

% \end{document}

\end{document}